\documentclass[letterpaper]{article} 
\usepackage{aaai25}  
\usepackage{times}  
\usepackage{helvet}  
\usepackage{courier}  
\usepackage[hyphens]{url}  
\usepackage{graphicx} 
\usepackage{natbib}  
\usepackage{caption} 
\usepackage[utf8]{inputenc} 
\usepackage[T1]{fontenc}    
\usepackage{hyperref}       
\usepackage{url}            
\usepackage{booktabs}       
\usepackage{amsfonts}       
\usepackage{nicefrac}       
\usepackage{microtype}      
\usepackage{xcolor}         
\usepackage{amsmath}
\usepackage{amsthm}
\usepackage{amssymb}
\usepackage{graphicx}
\usepackage{multirow}
\usepackage{tikz}
\usetikzlibrary{positioning, shapes, arrows.meta, calc}
\usepackage{xcolor}
\usepackage{algorithm}
\usepackage{algpseudocode}

\newcommand{\h}{\textbf{h}}
\newcommand{\x}{\textbf{x}}
\newcommand{\vel}{\textbf{v}}
\newcommand{\m}{\textbf{m}}

\newcommand{\thetab}{\boldsymbol{\theta}}
\newcommand{\xib}{\boldsymbol{\xi}}
\newcommand{\omegab}{\boldsymbol{\omega}}
\newcommand{\R}{\mathbb{R}}
\newcommand{\query}{\textbf{q}}
\newcommand{\val}{\textbf{v}}
\newcommand{\key}{\textbf{k}}
\makeatletter
\newcommand{\vast}{\bBigg@{4}}
\newcommand{\Vast}{\bBigg@{5}}
\makeatother

\newcommand\isomto{\stackrel{\sim}{\smash{\longrightarrow}\rule{0pt}{0.4ex}}}

\theoremstyle{definition}
\newtheorem{definition}{Definition}[section]

\usepackage{newfloat}
\usepackage{listings}
\DeclareCaptionStyle{ruled}{labelfont=normalfont,labelsep=colon,strut=off} 
\floatstyle{ruled}
\newfloat{listing}{tb}{lst}{}
\floatname{listing}{Listing}
\nocopyright
\title{Spacetime $E(n)$-Transformer: Equivariant Attention for Spatio-temporal Graphs}
\author {
    Sergio G. Charles
}
\affiliations{
    Stanford University\\
    sergio@cs.stanford.edu
}

\usepackage{bibentry}

\begin{document}

\maketitle

\begin{abstract}
  We introduce an $E(n)$-equivariant Transformer architecture for spatio-temporal graph data. By imposing rotation, translation, and permutation equivariance inductive biases in both space and time, we show that the Spacetime $E(n)$-Transformer (SET) outperforms purely spatial and temporal models without symmetry-preserving properties. We benchmark SET against said models on the charged $N$-body problem, a simple physical system with complex dynamics. While existing spatio-temporal graph neural networks focus on sequential modeling, we empirically demonstrate that leveraging underlying domain symmetries yields considerable improvements for modeling dynamical systems on graphs.
\end{abstract}


\begin{links}
    \link{Code}{https://github.com/sergiogcharles/set.git}
\end{links}

\section{Introduction}

Many problems that we wish to model with neural networks possess underlying geometric structure with symmetries. \textit{Geometric Deep Learning}, a term coined in the seminal work of \cite{gdl}, is an Erlangen program for deep learning that systematizes inductive biases as group symmetries $G$, arising through notions of invariance and equivariance.

Recent work, like $SE(3)$-Transformers \cite{se3} and $E(n)$-Graph Neural Networks \cite{egnn}, impose different notions of group equivariance on neural networks to inform architecture choice. While these neural network architectures encode spatial inductive biases, they notably lack a time component. Temporal Graph Networks \cite{tgn} proposed an efficient framework that learns from dynamic graphs. However, this architecture assumes the topology of graphs changes over time. In this paper, we discuss \textit{spatio-temporal graphs}, which have a fixed topology with changing features over discrete time steps. Recent works \cite{ming}, \cite{marisca}, \cite{cini} have treated node features as time series and edges as the relationships between these series. Such spatio-temporal graph neural networks (STGNNs) have a plethora of applications, from simulating biomolecular interactions to modeling financial time series.

Similarly, while STGNNs improve representation learning of sequential graph data, to the best of our knowledge, minimal research has been done on preserving group symmetries in a spatio-temporal fashion. In particular, sequential models ought to preserve spatial group symmetries at each time step. Famously, Noether's first theorem formalizes the notion of infinitesimal symmetries of the so-called Lagrangian of a physical system, in terms of perturbations with respect to both space and time, by determining conserved quantities. Inspired by this intuition of temporal and spatial symmetry, we seek to derive a neural network architecture that is equivariant in both temporal and spatial components.

Classical neural network architectures like RNNs are discrete approximations to continuous time-domain signals, obeying a differential equation with respect to time. If an RNN is invariant to \textit{time-warping}, a monotonically increasing and differentiable function of time, it takes the form of an LSTM \cite{gdl}, which unlike a vanilla RNN, captures long-term dependencies. Similarly, the dynamics of classical physical systems satisfy the Euler-Lagrange equations, i.e. the equations of motion. Hence, we use the charged $N$-body problem, as described in \cite{egnn} and adapted from \cite{nbody}, as an ideal candidate to test our hypothesis that preserving \textit{$G$-equivariance ameloriates long-term spatio-temporal graph modeling}. We will use a Transformer for the temporal component of the architecture, preserving long-term dependencies and, hence, invariance to time-warping. Each node, representing a charged particle of the graph, will have features, coordinates, and velocities. As such, the neural network should be equivariant under rotational and translational symmetries $E(n)$ acting on coordinates. It should also be equivariant with respect to rotational symmetries $SO(n)$ acting on velocities. Lastly, the nodes should be permutation equivariant.

\section{Background}

\subsection{Geometric Deep Learning}

Following the insights of Geometric Deep Learning \cite{gdl}, the input signals to machine learning models have an underlying domain $\Omega$. Examples of such domains include grids, graphs, and manifolds. The space of signals over $\Omega$ possesses a vector-space structure. That is \cite{gdl}:

\begin{definition}
    The space of $\mathcal{C}$-valued signals on $\Omega$ is 
    \begin{equation*}
        \mathcal{X}(\Omega, \mathcal{C}) = \{x:\Omega\to \mathcal{C}\},
    \end{equation*}
    which is a vector space of functions.
\end{definition}

The symmetry of the domain $\Omega$ will impose structure on the signal space $\mathcal{X}(\Omega)$, thus inducing structure on the space of interpolants 
\begin{equation*}
    \mathcal{F}(\mathcal{X}(\Omega))=\{f_{\theta\in\Theta}\}
\end{equation*}
for $f_{\theta}$ a neural network. In what follows, we canonically refer to $\mathcal{X}(\Omega)$ as $V$ for brevity.

\subsection{Group Representations, Invariance, and Equivariance}

\begin{definition}
    A \textit{representation} of a group $G$ on a vectorspace $V$ over a field $K$ is a homomorphism 
\begin{equation*}
    \rho:G\to GL(K, V)
\end{equation*}
such that $\rho(gh)=\rho(g)\rho(h)$ for all $g\in G$, $h\in G$, where $GL(K, V)$ is the general linear group of automorphisms $\varphi:V\isomto V$, i.e., the set of bijective linear transformations with function composition as its binary operation.
\end{definition}

In this paper, we are interested in the group of rotational symmetries $SO(n)$ and the group of isometries $E(n)$ of $\mathbb{R}^n$, as these are the naturally-induced symmetries of particles. Rotations are distance, angle, and orientation preserving transformations. The group of rotations in $n$ dimensions is
\begin{equation*}
    SO(n)=\{Q\in M_n(\mathbb{R}) | Q^\top Q = I \text{ and } \det Q = +1\},
\end{equation*}
where $M_n(\mathbb{R})$ is the set of $n\times n$ matrices with entries in $\mathbb{R}$, forming a ring. We represent a group element $g\in SO(n)$ with $\rho(g)\in GL(\R, \R^n)$, acting on $\x\in\R^n$ as $\rho(g):\x\mapsto Q\x$ where $Q\in\R^{n\times n}$ is an orthogonal matrix (see Appendix \ref{A} for more details). We restrict the notion of equivariance to functions of Euclidean space, as will be the case for neural networks. In Appendix \ref{A}, we provide a more general definition.

\begin{definition}
A function $f:\R^n\to \R^n$ is $SO(n)$-equivariant if 
\begin{equation*}
    Qf(\mathbf{x})=f(Q\mathbf{x})
\end{equation*}
for all $Q\in\mathbb{R}^{n\times n}$ orthogonal and $\mathbf{x}\in\R^n$.
\end{definition}

The Euclidean group $E(n)$ is the set of isometries of Eucliden space $\mathbb{R}^n$, i.e. transformations that preserve distance between points, represented as a rotation followed by a translation. More precisely, $E(n)=\{\varphi:\R^n\to\R^n|\varphi \text{ isometry}\}$. We represent a group element $g\in E(n)$ with $\rho(g)\in GL(\mathbb{R}, \mathbb{R}^n)$, acting on $\x\in\R^n$ as $\rho(g):\x\mapsto Q\x + \mathbf{b}$ where $Q\in\mathbb{R}^{n\times n}$ is an orthogonal rotation matrix and $\mathbf{b}\in\mathbb{R}^n$ is a translation vector. That is, we represent actions of the Euclidean group as orthogonal transformations followed by translations. 
Again, we provide a definition of equivariance with respect to functions of Euclidean space.

\begin{definition}
A function $f:\R^n\to\R^n$ is $E(n)$-equivariant if 
    \begin{equation*}        Qf(\mathbf{x})+\mathbf{b}=f(Q\mathbf{x}+\mathbf{b})
    \end{equation*}
for all $Q\in\mathbb{R}^{n\times n}$ orthogonal rotation matrices, $\mathbf{b}\in\mathbb{R}^n$ translation vectors, and for all $\mathbf{x}\in\mathbb{R}^n$.
\end{definition}
\section{Method}

In this paper, we are interested in physical systems that can be modelled as a sequence of graphs $G_t = (\mathcal{V}_t, \mathcal{E}_t)$ for $t=1,\dots, L$ with nodes $v_i(t)\in \mathcal{V}_t$ and edges $e_{ij}(t)\in\mathcal{E}_t$. In particular, we seek to model the dynamics of the charged $N$-body problem \cite{nbody}. For this task, we assume a priori that the graph is complete since a charged particle will interact with every other particle in a Van der Waals potential under Coulomb's law. In addition, we assume that particles are neither created nor destroyed as the system evolves in time, so the nodes $\mathcal{V}_t$ in the graph remain the same. Let $\mathcal{G}:=(G_t)_{1\le t\le L}$ be a sequence of topologically-identical graphs with changing features, known as a \textit{spatio-temporal graph}. The task under consideration is learning a function that predicts the associated features of graph. In particular, given $\mathcal{G}$, we are interested in predicting the positions and velocities of all particles in the system after $H$ additional time steps where $H>>L$.

To equip the spatio-temporal model of particle interactions with the appropriate inductive biases, we leverage both spatial and temporal notions of attention. For node $i$ at time step $t$ to attend to all the past neighborhoods of that node, we need to (1) aggregate nodes spatially to obtain spatially-contextual embeddings and (2) obtain temporally-contextual embeddings via temporal aggregation.

We fix a time slice $t$ such that the features derive from $G_t=(\mathcal{V}_t, \mathcal{E}_t)$. From the current features $\textbf{h}_i^{(l)}(t)$ of node $i$ at layer $l$, we form the next layer features $\textbf{h}_i^{(l+1)}(t)$ by aggregating neighboring node features. In particular,

\begin{equation}
    \textbf{h}_i^{(l+1)}(t) = \phi\left(\textbf{h}_i^{(l)}(t), \bigoplus_{j\in \mathcal{N}_i} a(\textbf{h}^{(l)}_i(t), \textbf{h}^{(l)}_j(t)) \psi(\textbf{h}^{(l)}_j(t))\right),
\end{equation}

where $\oplus$ is a permutation-invariant function \cite{gdl}, and $a$ is a self-attention mechanism, often a normalized softmax across neighbors. 
\subsection{$E(n)$-Equivariant Spatial Attention}\label{egnn}

\cite{egnn} introduced $E(n)$-Equivariant Graph Neural Networks (EGNNs). Every node in the graph $G=(\mathcal{V}, \mathcal{E})$ has features $\h_i\in\mathbb{R}^d$ and coordinates $\x_i\in\R^n$. In addition, we keep track of each particle's velocity $\vel_i\in\mathbb{R}^n$. The Equivariant Graph Convolutional Layer (EGCL) takes the set of node embeddings $h^{(l)}=\left\{\h_1^{(l)}, \dots, \h_N^{(l)}\right\}$, coordinate embeddings $x^{(l)}=\left\{\x_1^{(l)}, \dots, \x_N^{(l)}\right\}$, velocity embeddings $v^{(l)}=\left\{\boldsymbol{\vel}_1^{(l)}, \dots, \boldsymbol{\vel}_N^{(l)}\right\}$, and edge information $\mathcal{E}=(e_{ij})$ as input and produces the embeddings of the next layer. That is, $h^{(l+1)}, x^{(l+1)}, v^{(l+1)}=\text{EGCL}[h^{(l)}, x^{(l)}, v^{(l)}, \mathcal{E}]$, defined as follows \cite{egnn}:

\begin{equation}
\begin{split}
    \m_{ij}&=\phi_e\left(\h_i^{(l)}, \h_j^{(l)}, ||\x_i^{(l)}-\x_j^{(l)}||^2, a_{ij}\right)\\
    \vel_i^{(l+1)}&=\phi_v(\h_i^{(l)})\vel_i^{(l)} + C\sum_{j\ne i} (\x_i^{(l)}-\x_j^{(l)})\phi_x(\m_{ij})\\
    \x_i^{(l+1)}&=\x_i^{(l)} + \vel_i^{(l+1)}\\
    \m_i&=\sum_{j\ne i} \m_{ij}\\
    \h_i^{(l+1)}&=\phi_h(\h_i^{(l)}, \m_i)
\end{split}
\end{equation}

where $a_{ij}$ are the edge attributes, e.g. the edge values $e_{ij}$, and $\phi_e:\mathbb{R}^{2d+2}\to \mathbb{\mathbb{R}}^h$, $\phi_v:\mathbb{R}^d\to\mathbb{R}$, $\phi_x:\mathbb{R}^h\to\mathbb{R}$, and $\phi_h:\mathbb{R}^{d+h}\to\mathbb{R}^{d'}$ are MLPs. In what follows, we assume $d'=d$ for clarity.

\cite{egnn} proved that this layer is equivariant to rotations and translations on coordinates and equivariant to rotations on velocities:
\begin{equation}\small
    \h_i^{(l+1)}, Q\x_i^{(l+1)} + \mathbf{b}, Q\vel_i^{(l+1)} = \text{EGCL}[\h_i^{(l)}, Q\x_i^{(l)} + \mathbf{b}, Q\vel_i^{(l)}, \mathcal{E}]
\end{equation}

for $Q\in\mathbb{R}^{n\times n}$ an orthogonal rotation matrix and $\textbf{b}\in\mathbb{R}^n$ a translation vector. The EGCL is also permutation equivariant with respect to nodes $\mathcal{V}$.

Since we have a sequence of graphs $\mathcal{G}=\{G_t\}_{1\le t\le L}$, for a time slice $t$, we apply $K_t$ such EGCL transformation layers to the graph $G_t\in \mathcal{G}$:
\begin{equation}\small
    \h_i^{(l+1)}(t), \x_i^{(l+1)}(t), \vel_i^{(l+1)}(t)=\text{EGCL}[\h_i^{(l)}(t), \x_i^{(l)}(t), \vel_i^{(l)}(t), \mathcal{E}(t)]
\end{equation} 
for $l=1,\dots, K_t$. Thus, we obtain \textit{spatially-contextual} representations for node $i$ at time $t$ defined as $\boldsymbol{\theta}_i(t) = \h_i^{(K_t)}(t) \in\R^d$, $\boldsymbol{\xi}_i(t)=\x_i^{(K_t)}(t)\in\R^n$, $\omegab_i(t)=\vel_i^{(K_t)}(t)\in\R^n$ for $t=1,\dots, L$.

\subsection{Temporal Attention for Graphs}

The objective of this section is to obtain strong \textit{temporally-contextual} representations of the spatial graph embeddings. In the charged $N$-body problem, we are essentially solving the forward-time Euler-Lagrange equations, a second-order partial differential equation. However, for a fixed node on the spatio-temporal graph, the feature, position, and velocity form a time-series, for which RNN's capture short-term dependencies. It was shown in \cite{warp} that while vanilla RNNs are not time-warping invariant, LSTMs are a class of such time-warping invariant functions modeling a continuous time-domain signal. Employing this philosophy, the use of an attention-based Transformer architecture to model spatio-temporal graph data merits investigation.

\subsection{$E(n)$-Equivariant Temporal Attention}\label{sec:etal}

We would like the temporal attention to retain the equivariant properties described in Section \ref{egnn}. Namely, the Equivariant Temporal Attention Layer (ETAL) should be equivariant to the actions of $E(n)$ on coordinates and the actions of $SO(n)$ on velocities. It should also be permutation equivariant with respect to the actions of the symmetric group $\Sigma_N$ on nodes.

As we do not impose symmetry conditions on the layer with respect to the feature representations $\h_i(t)$, we can apply key-query-value self-attention as follows. Define the node-wise query $\query_i(t)=Q_i\thetab_i(t)$, key $\key_i(t)=K_i\thetab_i(t)$, and value $\val_i(t)=V_i\thetab_i(t)$ for $Q_i, K_i, V_i\in\mathbb{R}^{d\times d}$. To reduce memory usage, we share $Q, K, V$ for all nodes. Then the temporally-contextual representation is:

\begin{equation}\label{feature_update}
    \tilde{\thetab}_i(t):=\sum_{s=1}^L \alpha_i(t,s)\val_i(s)
\end{equation}

where
\begin{equation}
    \alpha_i(t,s)=\frac{\exp(\query_i(t)^\top \key_i(s))}{\sum_{s'=1}^L\exp(\query_i(t)^\top \key_i(s'))}
\end{equation}

Satorras et al. \cite{egnn} showed that for a collection of points $\{\boldsymbol{\xi_i}\}_{i=1}^N \in\R^{n}$, the norm is a unique geometric identifier, such that collections separated by actions of $E(n)$ form an equivalence class. With this in mind, since we desire the attention mechanism for coordinates $\xib_i(t)$ to be equivariant with respect to $E(n)$, we can define:

\begin{equation}\label{position_update}
\tilde{\xib}_i(t):=\xib_i(t)+B\sum_{\substack{s=1 \\ s \ne t}}^L \beta_i(t,s)(\xib_i(s)-\xib_i(t)),
\end{equation} where

\begin{equation}\label{beta}
    \beta_i(t,s)=\frac{\exp(||\xib_i(t)-\xib_i(s)||^2)}{\sum_{s'=1}^L\exp(||\xib_i(t)-\xib_i(s')||^2)}
\end{equation}

This bears semblance to the neighborhood attention described in the $SE(3)$-Transformer network \cite{se3} and Tensor Field Network layer \cite{tfn}, the intensity function in \cite{zhang}, and the invariant point attention in \cite{alphafold}.

We define an $SO(n)$-equivariant attention layer for velocities $\omegab_i(t)$:

\begin{equation}\label{velocity_update}
    \tilde{\omegab}_i(t):=\sum_{s=1}^L \gamma_i(t,s)\omegab_i(s)
\end{equation}
where the weight is
\begin{equation}
    \gamma_i(t,s)=\frac{\omegab_i(t)^\top \omegab_i(s)}{\sum_{s'=1}^L \exp(\omegab_i(t)^\top \omegab_i(s'))}.
\end{equation}

\begin{algorithm*}[t]
\caption{Spatiotemporal Attention (SpatiotempAttn)}\label{alg:attention}
\begin{algorithmic}
\Require $h$, $x$, $v$, $A$
\Comment{$h\in\R^{L\times N\times d}$, $x, v\in\R^{L\times N\times n}$, $A\in\R^{L\times N\times N}$}
\Require $E:\R^{N\times N}\times \R^{N\times n}\to \R^{N\times (N - 1)\times 2}$
\Require $W^{[1:L]}$, $X^{[1:L]}$, $Y^{[1:L]}$, $Z^{[1:L]}$
\Comment{$W^{[1:L]}\in \R^{L\times N\times d}$, $X^{[1:L]}\in \R^{L\times N\times n}$, $Y^{[1:L]}\in\R^{L\times N\times n}$, $Z^{[1:L]}\in\R^{L\times N\times N}$}

\State Initialize MLPs $f_{\theta}:\R^{L\times N\times d}\to\R^{L\times N\times d}$, $f_{A}:\R^{L\times N\times N}\to\R^{L\times N\times N}$

\For{$t=1,\dots, L$}
\State \Comment{Equivariant Spatial Attention Layer}
\State $h^{(1)}(t)\gets h(t)$ \Comment{$h^{(1)}(t)\in\R^{N\times d}$}
\State $x^{(1)}(t)\gets x(t)$ \Comment{$x^{(1)}(t)\in\R^{N\times n}$}
\State $v^{(1)}(t)\gets v(t)$ \Comment{$v^{(1)}(t)\in\R^{N\times n}$}
\State $\mathcal{E}(t)\gets E(A(t), x^{(1)}(t))$ \Comment{$\mathcal{E}(t)\in\R^{N\times (N-1)\times 2}$}

\For{$\ell=1,\dots, K - 1$}
\State $h^{(\ell+1)}(t), x^{(\ell + 1)}(t), v^{(\ell + 1)}(t)=\text{EGCL}[h^{(\ell)}(t), x^{(\ell)}(t), v^{(\ell)}(t),\mathcal{E}(t)]$
\EndFor
\State $\theta(t)\gets h^{(K)}(t)$ \Comment{$\theta(t)\in\R^{N\times d}$}
\State $\xi(t)\gets x^{(K)}(t)$ \Comment{$\xi(t)\in\R^{N\times n}$}
\State $\omega(t)\gets v^{(K)}(t)$ \Comment{$\omega(t)\in\R^{N\times n}$}
\EndFor

\State\Comment{Equivariant Temporal Attention Layer}

\State $\theta^{[1:L]}\gets (\theta(1),\dots,\theta(L))$ \Comment{$\theta^{[1:L]}\in\R^{L\times N\times d}$}

\State $\xi^{[1:L]}\gets (\xi(1),\dots,\xi(L))$ \Comment{$\xi^{[1:L]}\in\R^{L\times N\times n}$}

\State $\omega^{[1:L]}\gets (\omega(1),\dots,\omega(L))$ \Comment{$\omega^{[1:L]}\in\R^{L\times N\times n}$}

\State $A^{[1:L]}\gets A$ \Comment{$A^{[1:L]}\in\R^{L\times N\times N}$}
\State
\State $\hat{\theta}^{[1:L]}, \tilde{\xi}^{[1:L]}, \tilde{\omega}^{[1:L]}, \hat{A}^{[1:L]}=\text{ETAL}\left[\theta^{[1:L]} + W^{[1:L]}, \xi^{[1:L]}+X^{[1:L]}, \omega^{[1:L]}+Y^{[1:L]},
        A^{[1:L]}+Z^{[1:L]}\right]$
\State $\tilde{\theta}^{[1:L]}=f_{\theta}(\text{LN}(\hat{\theta}^{[1:L]}))+\hat{\theta}^{[1:L]}$

\State $\tilde{A}^{[1:L]}=f_{A}(\text{LN}(\hat{A}^{[1:L]}))+\hat{A}^{[1:L]}$

\State \Return $\tilde{\theta}^{[1:L]}, \tilde{\xi}^{[1:L]}, \tilde{\omega}^{[1:L]}, \tilde{A}^{[1:L]}$
\end{algorithmic}
\end{algorithm*}

In appendix \ref{B}, we show that the position attention function is $E(n)$-equivariant and the velocity attention function is $SO(n)$-equivariant. 

Following the insights of \cite{ming}, edges are relationships between time series and they should evolve. Hence, while the adjacency matrix $A\in\R^{N\times N}$ is constant in space when applying ECGL, it should intuitively evolve in time when applying ETAL. That is, if we consider edges as representing the interaction between charged particles, e.g. the strength of the force, then this must necessarily evolve in time for a non-stationary point cloud system.

We define a key matrix $K(t)=KA(t) \in \R^{N\times N}$, query matrix $Q(t)=QA(t) \in \R^{N\times N}$, and value matrix $V(t)=VA(t) \in \R^{N\times N}$ for $t=1,\dots, L$ and $K,Q, V\in\R^{N\times N}$. Thus, to obtain a temporally-contextual representation of the adjacency matrix at time $t$, we apply attention:
\begin{equation}
    \tilde{A}(t)=\sum_{s=1}^L \pi(t,s)V(s) \in\R^{N\times N}
\end{equation}
where
\begin{equation}
    \pi(t,s) = \exp(Q(t)^\top K(s))\left(\sum_{s'=1}^L \exp(Q(t)^\top K(s'))\right)^{-1}.
\end{equation}

Furthermore, in Appendix \ref{C}, we tensorize the feature, position, velocity, and adjacency components of ETAL to efficiently compute these operations in both space $i=1,\dots, N$ and time $t=1,\dots, L$ dimensions.

\subsection{Spacetime $E(n)$-Equivariant Graph Transformer}

The full spatio-temporal attention module is presented in Algorithm \ref{alg:attention}. It takes as input the node features $h\in\R^{L\times N\times d}$, positions $x\in\R^{L\times N\times n}$, velocities $v\in\R^{L\times N\times n}$ and adjacency matrices $A\in\R^{L\times N\times N}$. For a spatio-temporal graph $\mathcal{G}=(G_t)_{1\le t\le L}$, we apply an equivariant spatial attention layer in the form of EGCL to obtain spatially-contextual representations $\theta(t)\in\R^{N\times d}, \xi(t)\in\R^{N\times n}, \omega(t)\in\R^{N\times n}$ for $t=1,\dots, L$. We share the same EGCL layer across all time steps $t=1,\dots, L$. That is, we only learn one set of MLPs $\phi_e$, $\phi_v$, $\phi_x$, and $\phi_h$ for each layer across time, which is significantly more memory and parameter efficient. 

Observe, at each time step, we apply a transformation $E:\R^{N\times N}\times \R^{N\times n}\to \R^{N\times (N-1)\times 2}$ to the adjacency matrix $A(t)\in\R^{N\times N}$ and the coordinates $x(t)\in\R^{N\times n}$ for $G_t$. This will produce edge attributes $e_{ij}(t)=\left(c_i c_j, ||\x_i(t)-\x_j(t)||_2^2\right)$ containing charge and distance information for neighboring nodes. Since each graph is complete, there are $N\times (N-1)$ such edge attributes, which we store in the tensor $\mathcal{E}(t)\in \R^{N\times(N-1)\times 2}$.

\begin{algorithm*}[t]
\caption{Spacetime $E(n)$-Transformer (SET)}\label{alg:set}
\begin{algorithmic}
\Require $h$, $x$, $v$, $A$ \Comment{$h\in\R^{L\times N\times d}$, $x, v\in\R^{L\times N\times n}$, $A\in\R^{N\times N}$}
\State $\hat{\theta}_{(0)}^{[1:L]} \gets h$
\State $\hat{\xi}_{(0)}^{[1:L]} \gets x$
\State $\hat{\omega}_{(0)}^{[1:L]} \gets v$
\State $\hat{A}_{(0)}^{[1:L]} \gets (A,\dots, A)$\Comment{$\hat{A}_{(0)}^{[1:L]}\in\R^{L\times N\times N}$}

\For{$m=1,\dots, M$}
    \State $\hat{\theta}_{(m+1)}^{[1:L]}, \hat{\xi}_{(m+1)}^{[1:L]}, \hat{\omega}_{(m+1)}^{[1:L]}, \hat{A}_{(m+1)}^{[1:L]}\gets \text{SpatiotempAttn}\left(\hat{\theta}_{(m)}^{[1:L]}, \hat{\xi}_{(m)}^{[1:L]}, \hat{\omega}_{(m)}^{[1:L]}, \hat{A}_{(m)}^{[1:L]}\right)$
\EndFor
\State $\hat{x}(L+H)=\frac{1}{L}\sum_{t=1}^L \hat{\xi}_{(M)}(t)$ \Comment{$\hat{x}(L+H)\in\R^{N\times n}$}
\State $\hat{v}(L+H)=\frac{1}{L}\sum_{t=1}^L \hat{\omega}_{(M)}(t)$ \Comment{$\hat{x}(L+H)\in\R^{N\times n}$}

\State \Return $\hat{x}(L+H), \hat{v}(L+H)$
\end{algorithmic}
\end{algorithm*}

Then we apply equivariant temporal attention in the form of ETAL to the spatial representations $\theta^{[1:L]}\in\R^{L\times N\times d}$, $\xi^{[1:L]}\in\R^{L\times N\times n}$, and $\omega^{[1:L]}\in\R^{L\times N\times n}$. Feed-forward networks $f_{\theta}:\R^{L\times N\times d}\to\R^{L\times N\times d}$, $f_{A}:\R^{L\times N\times N}\to \R^{L\times N\times N}$ with layer pre-normalization, defined in Appendix \ref{D}, and residual connection are also applied to the respective feature and edge components of the graph. The sinusoidal positional encodings $W^{[1:L]}\in\R^{L\times N\times d}$, $X^{[1:L]}\in\R^{L\times n\times n}$, $Y^{[1:L]}\in\R^{L\times N\times n}$, $Z^{[1:L]}\in\R^{L\times N\times N}$ for the features, positions, velocities, and adjacency matrices are defined in Appendix \ref{D}.

\begin{figure}[H] 
    \centering
\resizebox{0.5\textwidth}{!}{ 
\begin{tikzpicture}
    \tikzstyle{block} = [rectangle, draw, text centered, minimum height=2em]
    \tikzstyle{arrow} = [thick,->,>=stealth]
    \tikzstyle{element} = [rectangle, draw, text centered, minimum height=2.4em, minimum width=6em]
    \tikzstyle{longblock} = [rectangle, draw, text centered, minimum height=2.4em, minimum width=44em]
    \tikzstyle{transparentblock} = [rectangle, text centered, minimum height=2.4em, fill=none]

    \definecolor{lightyellow}{HTML}{FFFFE0}
    \definecolor{lightgreen}{HTML}{90EE90}
    \definecolor{lightpurple}{HTML}{E6E6FA}
    \definecolor{lightorange}{HTML}{FFDAB9}
    \definecolor{lightred}{HTML}{FFCCCB}
    \definecolor{lightblue}{HTML}{edf5ff}
    \definecolor{lightgrey}{HTML}{D3D3D3}
    \definecolor{darkyellow}{HTML}{FFD700}
    \definecolor{darkgreen}{HTML}{228B22}
    \definecolor{darkpurple}{HTML}{9391ff}
    \definecolor{darkorange}{HTML}{FF8C00}
    \definecolor{darkred}{HTML}{8B0000}
    
    \tikzstyle{colorblock} = [rectangle, draw, fill=lightblue, text centered, minimum height=15.5cm, minimum width=19cm, opacity=0.5]

    \tikzstyle{noborder} = [rectangle, text centered, minimum height=2.4em, minimum width=6em]
    
    \node (input) at (0, 6) {\includegraphics[width=3.0cm, height=3.0cm]{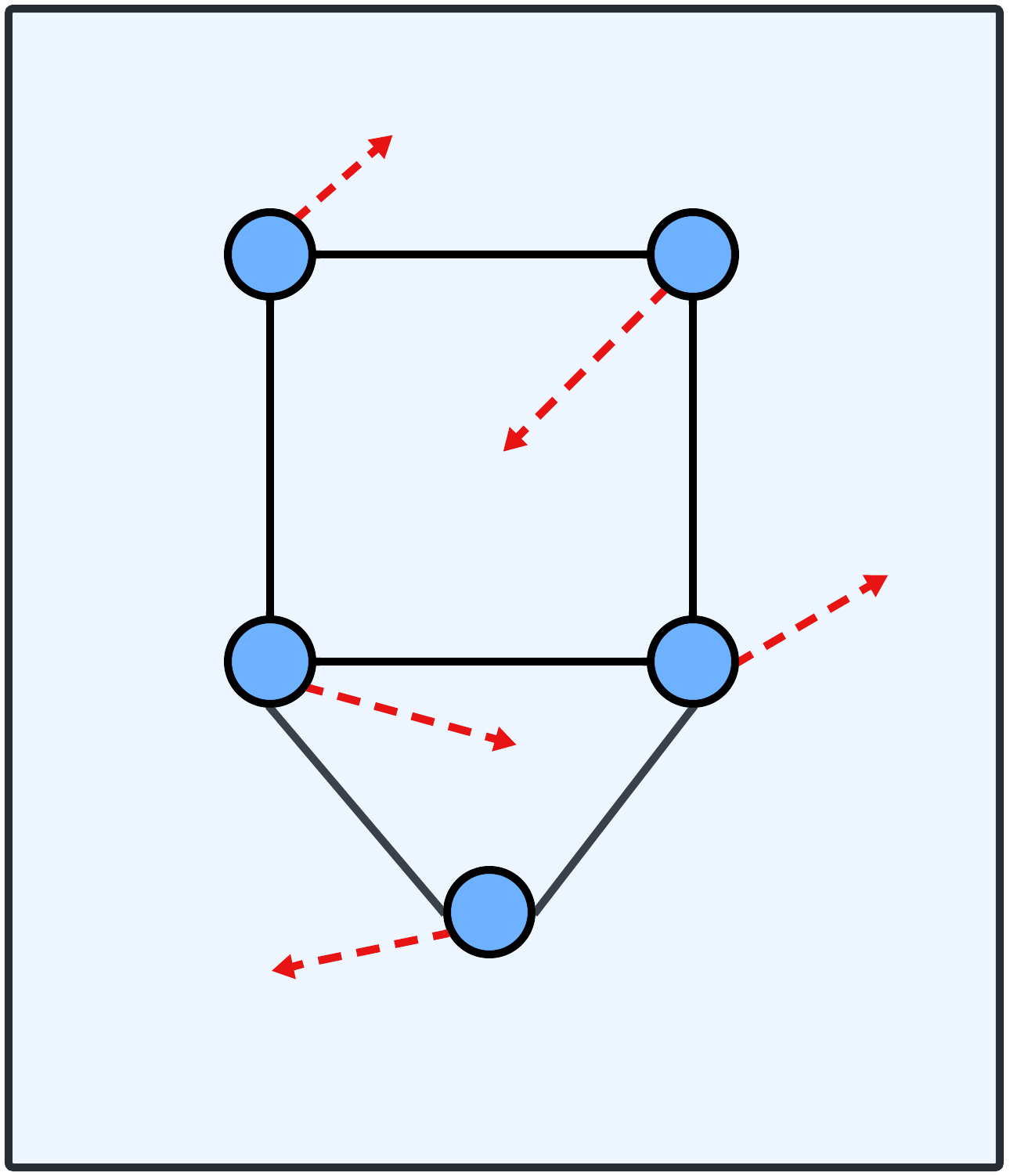}};
    \node[block, fill=lightgrey] (mean) at (0, 3) {Mean};

    \node[colorblock] (coloredblock) at (0, -5.5) {};
    \node[noborder] (repeat) at (10.25, -5) {$M\times$};

    \node[element] (left-input) at (-4.0, 1.5) {$(\theta_{(m)}(1),A_{(m)}(1),\xi_{(m)}(1), \omega_{(m)}(1))$};
    \node[element] (right-input) at (4.0, 1.5) {$(\theta_{(m)}(L),A_{(m)}(L),\xi_{(m)}(L), \omega_{(m)}(L))$};

    \node[block, fill=lightred] (concat-left) at (-4.0, 0) {Concat};
    \node[block, fill=lightred] (concat-right) at (4.0, 0) {Concat};

    \node[block, fill=lightorange] (addnorm-left) at (-4.0, -1.5) {Add+Norm};
    \node[block, fill=lightorange] (addnorm-right) at (4.0, -1.5) {Add+Norm};

    \node[block, fill=darkpurple] (fnn-left) at (-4.0, -3) {FNN};
    \node[block, fill=darkpurple] (fnn-right) at (4.0, -3) {FNN};

    \node[element] (left-xiomega) at (-7.052, -6) {$\left(\xi_{(m)}(1), \omega_{(m)}(L)\right)$};

    \node[element] (left-thetaA) at (-4.0, -6) {$\left(\tilde{\theta}_{(m)}(1), \hat{A}_{(m)}(1)\right)$};
    
    \node[element] (right-thetaA) at (3.95, -6) {$\left(\tilde{\theta}_{(m)}(L), \tilde{A}_{(m)}(L)\right)$};

    \node[element] (right-xiomega) at (7.09, -6) {$\left(\xi_{(m)}(L), \omega_{(m)}(L)\right)$};

    \node[transparentblock] (left_rep) at (-5.55, -6) {};
    \node[transparentblock] (right_rep) at (5.55, -6) {};

    \node[transparentblock] (left_xiomega_center) at (-7.0, -6) {};
    \node[transparentblock] (left_thetaA_center) at (-4.0, -6) {};
    \node[transparentblock] (left_thetaA_rightside) at (-2.7, -6) {};
    
    \node[transparentblock] (right_thetaA_center) at (4.0, -6) {};
    \node[transparentblock] (right_xiomega_center) at (7.0, -6) {};
    \node[transparentblock] (right_thetaA_leftside) at (2.6, -6) {};

    \node[transparentblock] (left_etal) at (-5.55, -8) {};
    \node[transparentblock] (right_etal) at (5.55, -8) {};

    \node[longblock, fill=lightgreen] (etal) at (0, -8) {ETAL};

    \node[transparentblock] (dots) at (0, 1.5) {$\cdots$};
    \node[transparentblock] (dots) at (0, -6) {$\cdots$};
    \node[transparentblock] (dots) at (0, -10) {$\cdots$};
    \node[transparentblock] (dots) at (0, -15) {$\cdots$};

    \node[below=of left_etal, draw, thick, circle] (embplus_left) {$+$};

    \node[left=of embplus_left, draw, thick, circle, inner sep=0pt,label={[align=left]left:Positional\\Encoding}] (pe_left) {\tikz \draw[scale=0.1] plot[domain=0.0:6.28] (\x,{sin(\x r)});};

    \node[below=of right_etal, draw, thick, circle] (embplus_right) {$+$};
    
    \node[right=of embplus_right, draw, thick, circle, inner sep=0pt,label={[align=right]right:Positional\\Encoding}] (pe_right) {\tikz \draw[scale=0.1] plot[domain=0.0:6.28] (\x,{sin(\x r)});};

    \node[block, fill=darkyellow] (egnn-left) at (-5.5, -11) {EGNN};
    \node[block, fill=darkyellow] (egnn-right) at (5.5, -11) {EGNN};

    \node[element] (left-bottom) at (-5.5, -12.5) {$\left(\xi_{(m)}(1), \omega_{(m)}(1)\right)$};
    \node[element] (right-bottom) at (5.5, -12.5) {$\left(\xi_{(m)}(L), \omega_{(m)}(L)\right)$};

    \node (left-bottom-image) at (-5.5, -15.5) {\includegraphics[width=3.0cm, height=3.0cm]{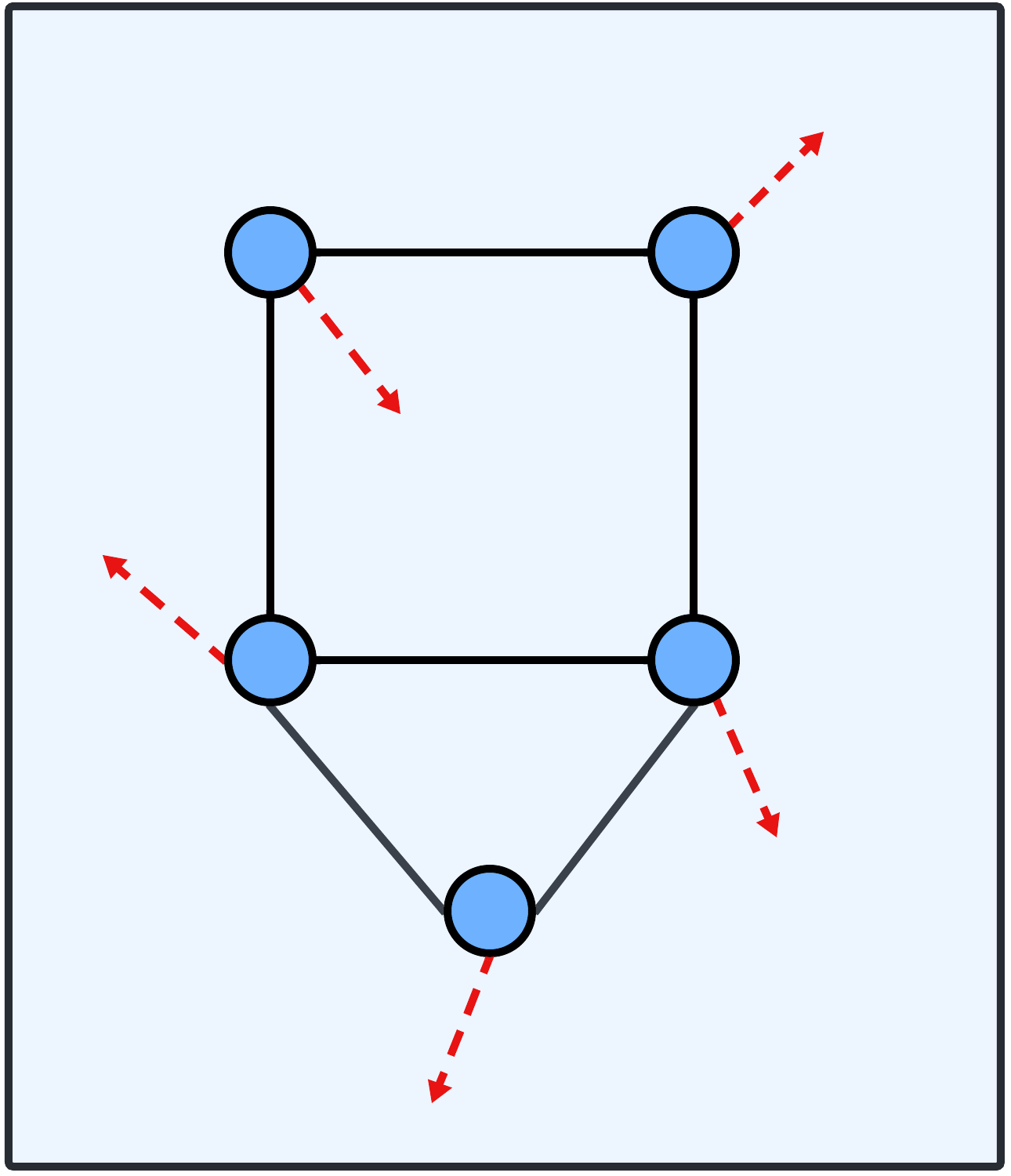}};
    \node (right-bottom-image) at (5.5, -15.5) {\includegraphics[width=3.0cm, height=3.0cm]{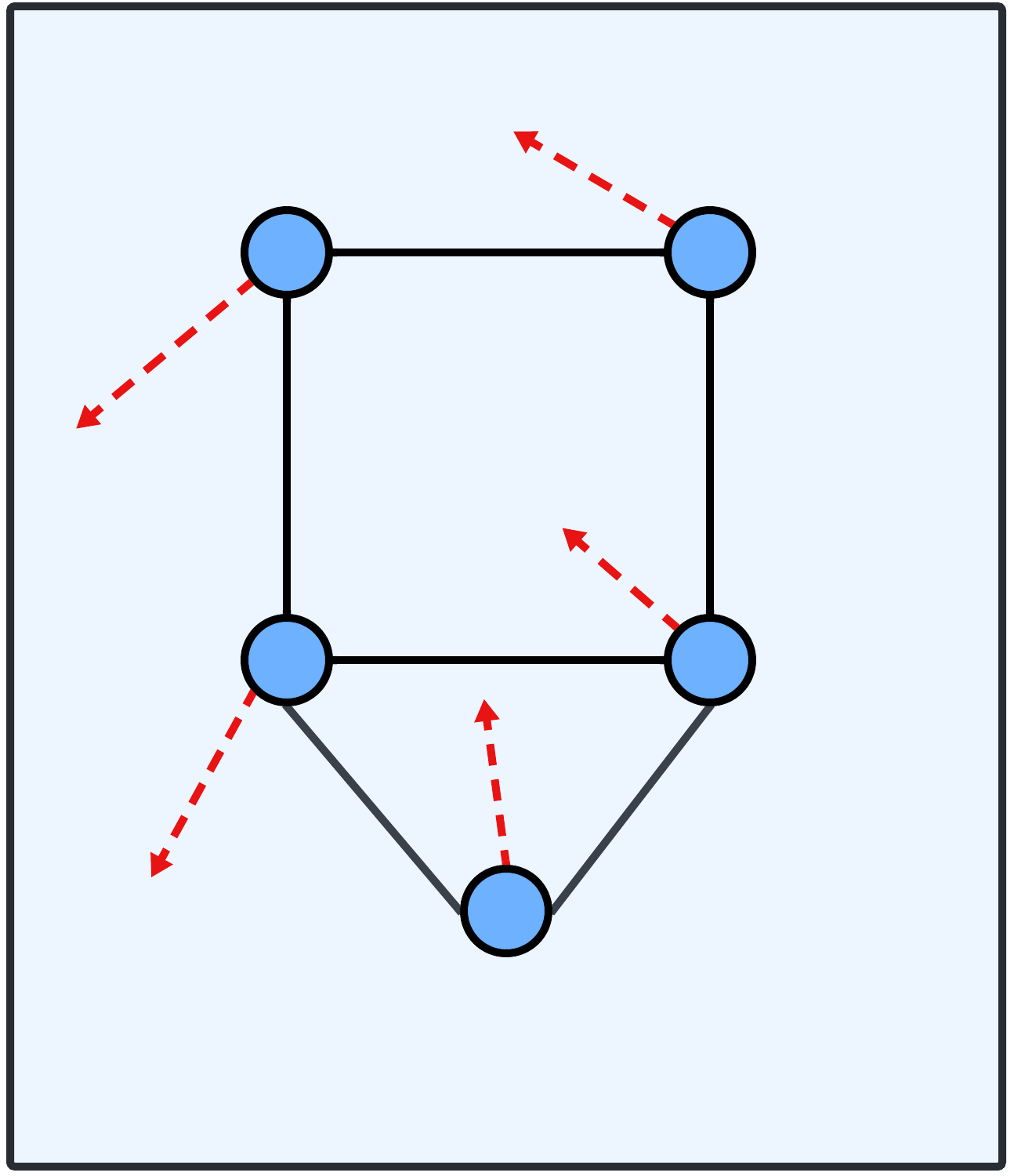}};

    \draw[arrow] (mean) -- (input);
    \draw[arrow] (left-input) -- (mean);
    \draw[arrow] (right-input) -- (mean);

    \draw[arrow] (concat-left) -- (left-input);
    \draw[arrow] (concat-right) -- (right-input);

    \draw[arrow] (addnorm-left) -- (concat-left);
    \draw[arrow] (addnorm-right) -- (concat-right);

    \draw[arrow] (fnn-left) -- (addnorm-left);
    \draw[arrow] (fnn-right) -- (addnorm-right);

    \draw[arrow] (left_thetaA_center) -- (fnn-left);
    \draw[arrow] (right_thetaA_center) -- (fnn-right);

    \draw[arrow] (left_xiomega_center) |- (concat-left);
    \draw[arrow] (right_xiomega_center) |- (concat-right);

    \draw[arrow] (left_thetaA_rightside) |- (addnorm-left);
    \draw[arrow] (right_thetaA_leftside) |- (addnorm-right);

    \draw[arrow] (left_etal) -- (left_rep);
    \draw[arrow] (right_etal) -- (right_rep);

    \draw[arrow] (pe_left)-- (embplus_left);
    \draw[arrow] (egnn-left) -- (embplus_left);
    \draw[arrow] (embplus_left) -- (left_etal);

    \draw[arrow] (pe_right) -- (embplus_right);
    \draw[arrow] (egnn-right) -- (embplus_right);
    \draw[arrow] (embplus_right) -- (right_etal);

    \draw[arrow] (left-bottom) -- (egnn-left);
    \draw[arrow] (right-bottom) -- (egnn-right);

    \draw[arrow] (left-bottom-image) -- (left-bottom);
    \draw[arrow] (right-bottom-image) -- (right-bottom);
\end{tikzpicture}
}
\caption{Spacetime $E(n)$-Transformer architecture.}
    \label{fig:arch}
\end{figure}
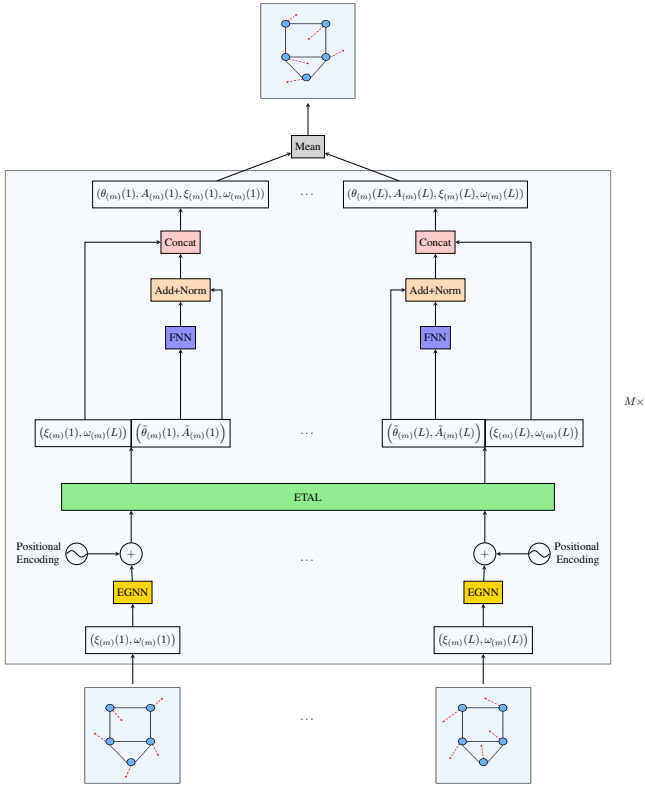

As the design of spatio-temporal attention is modular, we can continue stacking this architecture as we see fit (see Figure \ref{fig:arch}). In Algorithm \ref{alg:set}, we apply spatio-temporal attention $M$ times. Then we take a mean of the resulting \textit{spatio-temporally contextual} representations of positions and velocities across the time dimension, which we use as the predicted particles' coordinates $\hat{x}(L+H)\in\R^{N\times n}$ and velocities $\hat{v}(L+H)\in\R^{N\times n}$ at the horizon target $t=L+H$.

The task is to predict both the positions and velocities of particles at time $L+H$, so we minimize the following loss:

\begin{equation}\label{objective}
    \begin{split}
        \mathcal{L} &= ||\hat{x}(L+H)-x(L+H)||_2^2 + \alpha ||\hat{v}(L+H)-v(L+H)||_2^2\\
        &=\frac{1}{Nn}\sum_{i=1}^N\sum_{j=1}^n (\hat{x}_{ij}(L+H)-x_{ij}(L+H))^2\\
        &+\frac{\alpha}{Nn}\sum_{i=1}^N\sum_{j=1}^n (\hat{v}_{ij}(L+H)-v_{ij}(L+H))^2,
    \end{split}
\end{equation}
for $\alpha\in(0,1)$ a hyper-parameter.

\section{Related Work}

Temporal graph learning has a plethora of real-world applications, like COVID-19 contact tracing \cite{covid1} \cite{covid2} \cite{flight} and misinformation detection \cite{dyngcn} \cite{song} \cite{zhang}.

\begin{table*}[t]
\centering
\renewcommand{\arraystretch}{1.5}
\resizebox{\textwidth}{!}{
\begin{tabular}{llcccc}
\hline
\multicolumn{1}{c}{\textbf{Ablation}} & \multicolumn{1}{c}{\textbf{Model}} & \multicolumn{1}{c}{\textbf{Params}} & \multicolumn{1}{c}{\textbf{Val MSE}} & \multicolumn{1}{c}{\textbf{Test MSE}} & \multicolumn{1}{c}{\textbf{MSE Ratio}} \\
\hline
\multirow{2}{*}{Equivariance} & \textbf{Equiv=True}, Adj=False, SATT=True, TATT=True & 796,058 & \textbf{1.21e-10} & \textbf{1.25e-10} & $-$ \\
 & \textbf{Equiv=False}, Adj=False, SATT=True, TATT=True & 796,244 & 1.96e-10 & 2.03e-10 & 1.57$\times$ \\
\hline
Adjacency & Equiv=True, \textbf{Adj=True}, SATT=True, TATT=True & 810,458 & 1.12e-09 & 1.29e-10 & 8.96$\times$ \\
\hline
Attention & Equiv=True, Adj=False, \textbf{SATT=True}, \textbf{TATT=False} & 796,049 & 2.73e-10 & 3.57e-10 & 2.86$\times$ \\
\hline
\end{tabular}
}
\vspace{0.5cm}
\caption{Ablation study of equivariance, adjacency, and attention for $N=5$. We present the model settings, parameter counts, validation \& test MSE, and the MSE ratio, which is the ratio of the ablation model's test MSE divided by the best performing model's test MSE.}
\label{tab:ablation_study}
\end{table*}

Learning on continuous-time dynamic graphs was introduced by \cite{tgn}, which proposed Temporal Graph Networks (TGNs) with a memory module, acting as a summary of what the model has seen so far for each node. Causal Anonymous Walks \cite{caw} is another branch of temporal graph learning, which extracts random walks between edges; however, this is not our focus. Other work like \cite{ming} and \cite{cini} treat node features as time series and edges as correlations between the series. Under this framework, message passing must be able to handle \textit{sequences} of data from the neighborhood of each node, with RNNs \cite{seo}, attention mechanisms \cite{marisca}, and convolutions \cite{wu}.

The Dynamic Graph Convolutional Network (DynGCN) \cite{dyngcn} and DyGFormer \cite{dygformer} are most similar to our method. DynGCN processes each of the graph snapshots with a graph convolutional network to obtain \textit{structural information} and then applies an attention mechanism to capture \textit{temporal information}. Similarly, DyGFormer \cite{dygformer} learns from historical first-hop neighborhood interactions and applies a Transformer architecture to historical correlations between nodes. However, unlike our paper, DynGCN and DyGFormer do not take into account the inductive biases of the underlying modeling task.

$E(n)$-Equivariant Graph Neural Networks (EGNN) \cite{egnn} defines a model equivariant to the Euclidean group $E(n)$ and, unlike previous methods, does not rely on spherical harmonics such as the $SE(3)$-Transformer \cite{se3} and Tensor Field Networks \cite{tfn}. The $SE(3)$-Transformer paper briefly alludes to incorporating equivariant attention with an LSTM for temporal causality; however, this is not the primary focus of their work. LieConv \cite{lieconv} proposes a framework that allows one to construct a convolutional layer that is equivariant with respect to transformations of a Lie group, equipped with an exponential map. However, the EGNN is simpler and more applicable to problems with point clouds like the charged $N$-body problem \cite{nbody} we consider. 

As we concern ourselves with modeling a dynamical system, the works of Lagrangian Neural Networks (LNNs) \cite{lnn} and Hamiltonian Neural Networks (HNNs) \cite{hnn} are pertinent.  HNNs parameterize the Hamiltonian of a system, but require canonical coordinates, which makes it inapplicable to systems where such coordinates cannot be deduced. LLNs parameterize arbitrary Lagrangians of dynamical systems with neural networks, from which it is possible to solve the forward dynamics of the system; however, this requires an additional step of integration, which is cumbersome.

\section{Experiments \& Results}

Adapting the charged $N$-body system dataset from \cite{egnn}, we sample 16,000 trajectories for training, 2,000 trajectories for validation, and 2,000 trajectories for testing. Each trajectory has a horizon length of $H=10,000$ and a sequence length of $L=10$. The point cloud consists of $N=5$ particles, where at each time step, positions $(\x_1(t),\dots \x_5(t))^\top \in\R^{5\times 3}$, velocities $(\vel_1(t),\dots \vel_5(t))^\top \in\R^{5\times 3}$, as well as charges $c_1,\dots, c_5\in\{-1, +1\}$ are known. The edges between charged particles is $e_{ij}(t)=\left(c_i c_j, ||\x_i(t)-\x_j(t)||_2^2\right)$. We input these known values into SET with features chosen as $h_i(t)=||\vel_i(t)||_2$ for $i=1,\dots, 5$. We conducted a hyper-parameter optimization and selected the best model settings, as per Appendix \ref{E}.

\subsection{Ablation Study: Equivariance, Adjacency, and Attention}
Furthermore, we conduct an ablation study on SET, shown in Table \ref{tab:ablation_study}, which compares the use of equivariance, temporal attention for the adjacency matrix as per Section \ref{sec:etal}, spatial attention, and temporal attention. By selecting the best model on the validation set, we find that incorporating equivariance, spatial attention, and temporal attention enhances performance, whereas using adjacency diminishes it. We hypothesize that the insignificance of temporal adjacency is due to the fact the edge attribute contains information about charges, which does not evolve in time, and information about the distance between particles, which already implicitly exists in the coordinate information.
\subsection{Baselines \& Scaling $N$}
We compare our best performing SET model with optimized LSTM, EGNN, MLP and Linear baselines (see Appendix \ref{E} for implementation details). SET outperforms all baselines for $N=5$, as seen in Table \ref{tab:model_comparison}.

\begin{table}
\centering
\renewcommand{\arraystretch}{1.25}
\begin{tabular}{lcc}
\hline
\multicolumn{1}{l}{\textbf{Model}} & \multicolumn{1}{c}{\textbf{Params}} & \multicolumn{1}{c}{\textbf{Test MSE}} \\
\hline
SET & 796,058 & \textbf{1.25e-10} \\
LSTM & 826,313 & 2.03e-08 \\
EGNN & 100,612 & 2.05e-06 \\
MLP & 67,718 & 3.48e-06 \\
Linear & 3 & 3.04 \\
\hline
\end{tabular}
\vspace{0.5cm}
\caption{Baselines for $N=5$.}
\label{tab:model_comparison}
\end{table}

\begin{figure}[H]
\centering
\begin{minipage}{0.5\textwidth}
    \centering
    \includegraphics[width=\linewidth]{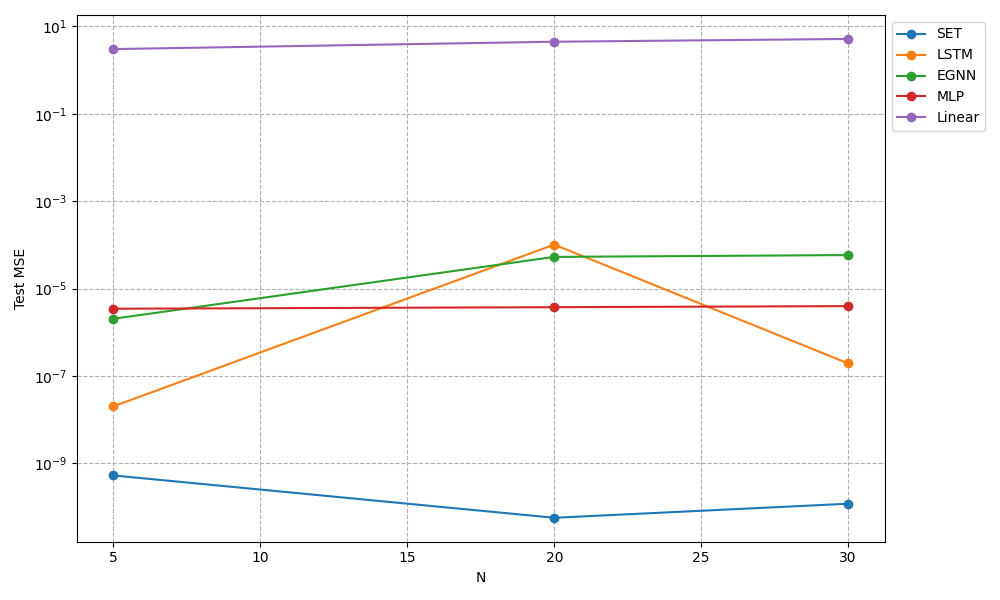}
\end{minipage}
\begin{minipage}{0.5\textwidth}
    \centering
    \includegraphics[width=\linewidth]{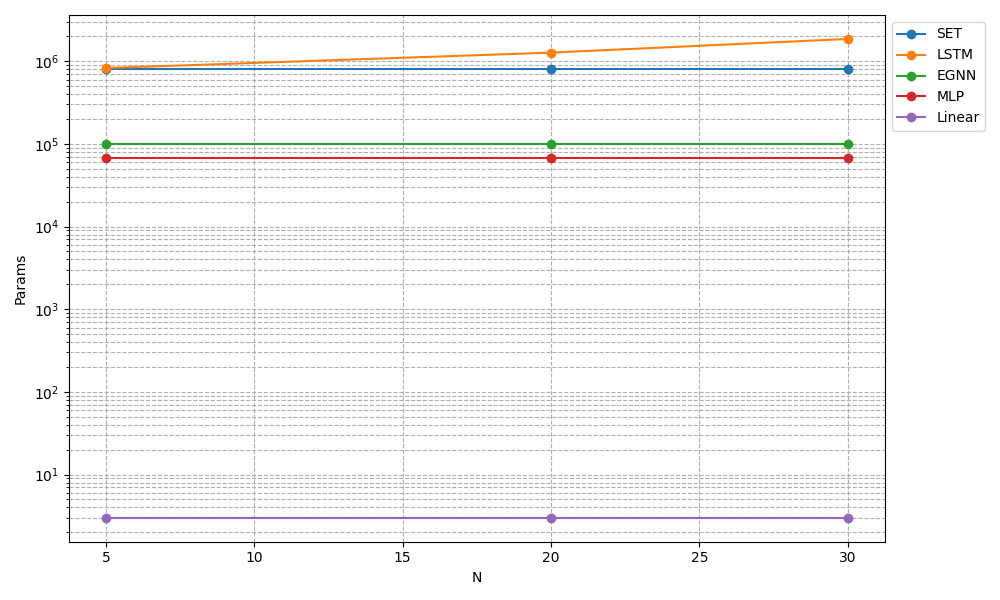}
\end{minipage}
\caption{\textbf{Top:} Model test MSE versus $N$. \textbf{Bottom:} Number of model parameters versus $N$.}
\label{fig:mse_and_params}
\end{figure}

Since the $N=5$ system is seemingly too simple a task, we scale the dataset to $N=20$ and $N=30$. As shown in Figure \ref{fig:mse_and_params}, test MSE remains consistent for all models regardless of the number of charged particles $N$, which is a desirable property. Per Figure \ref{fig:mse_and_params}, the number of model parameters remains constant for the EGNN, MLP and Linear baselines. Fortunately, the number of parameters in SET also remains constant for all $N$; this is an artifact of the attention layers only being functions of the feature and coordinate dimensions. Note, the number of parameters in the LSTM increases from $8.2e5$ to $1.8e6$, which is an undesirable property. Further results are included in Appendix \ref{E}. 

\subsection{Noisy Observations}
We also conduct an experiment in which we introduce noise in the observations, by adding Gaussian noise $\varepsilon\sim\mathcal{N}(0, 0.5)$ to positions and velocities. As seen in Table \ref{tab:model_comparison_noisy}, all models have test MSEs that plateau around the variance $\sigma^2=0.5$ of the irreducible noise, with the SET and MLP baseline tying in performance.

\begin{table}[H]
\centering
\renewcommand{\arraystretch}{1.25}
\begin{tabular}{lcc}
\hline
\multicolumn{1}{l}{\textbf{Model}} & \multicolumn{1}{c}{\textbf{Params}} & \multicolumn{1}{c}{\textbf{Test MSE}} \\
\hline
SET & 796,058 & \textbf{4.97e-01} \\
LSTM & 826,313 & 4.98e-01 \\
EGNN & 100,612 & 4.98e-01 \\
MLP & 67,718 & \textbf{4.97e-01} \\
Linear & 3 & 3.60 \\
\hline
\end{tabular}
\vspace{0.5cm}
\caption{Model comparison for a noisy $N=5$ dataset.}
\label{tab:model_comparison_noisy}
\end{table}

\section{Conclusions}

The imposition of group symmetries on graph neural networks is a promising area of research, demonstrating remarkable real-world results like AlphaFold \cite{alphafold}. However, most research has been centered on spatial equivariance for representational learning on static graphs. For dynamical graph systems, little research has centered on preserving group symmetries across time. We close this gap with the Spacetime $E(n)$-Transformer and show promising results for the charged $N$-body problem. It will be interesting to see our method applied to harder tasks, such as sequential bio-molecular generation.

Although we chose a graph as the domain of interest, it is plausible to extend notions of spatio-temporal $G$-equivariance to other domains like grids and manifolds. Furthermore, while we leveraged the symmetries of the problem a priori, it may not always be possible to find a simple group for a general problem. Hence, in future work, it would be interesting to learn a group symmetry from underlying data and impose equivariance using methods like LieConv \cite{lieconv}, which is equivariant to the actions of Lie groups, i.e. the continuous group representation of infinitesimal transformations. Noether's first theorem implies a possible connection to conserved quantities, which was discussed in \cite{noether}. 
\bibliography{aaai25}

\onecolumn
\appendix
\section{Preliminaries}\label{A}

\subsection{Isometries}
\begin{definition}
Let $(X, d_X)$ and $(Y, d_Y)$ be metric spaces with metrics $d_X:X\times X\to\mathbb{R}$, $d_Y:Y\times Y\to\mathbb{R}$. An isometry $\varphi:X\to Y$ is a distance-preserving isomorphism if
\begin{equation}
    d_X(a, b) = d_Y(\varphi(a), \varphi(b))
\end{equation}
for all $a, b\in X$. 
\end{definition}

\subsection{Invariance \& Equivariance}

\begin{definition}
Let $\rho_V:G\to GL(K,V)$ be a representation of group $G$, and let $\rho_V(g):V\isomto V$ be an automorphism on $V$ for $g\in G$. A function $f:V\to W$ is $G$-\textit{equivariant} if there exists an equivalent representation $\rho_W:G\to GL(K, W)$ with equivalent automorphism $\rho_W(g):W\isomto W$, such that
\begin{equation}
    f(\rho_V(g)(v)) = \rho_W(g)(f(v))
\end{equation}
for all $v\in V$ and $g\in G$.
\end{definition}

\begin{definition}
    Let $\rho:G\to GL(K,V)$ be a representation of group $G$ and let $\rho(g):V\isomto V$ be an automorphism for $g\in G$. A function $f:V\to W$ is $G$-\textit{invariant} if 
    \begin{equation}
        f(\rho(g)(v)) = f(v)
    \end{equation}
    for all $v\in V$ and $g\in G$.
\end{definition}

\subsection{Special Property of $SO(n)$}

Since rotations are distance, angle, and orientation preserving, they are linear transformations. As such, rotations can be represented as a matrix. Suppose $\mathbf{x},\mathbf{y}\in\mathbb{R}^n$ and $Q\in\mathbb{R}^{n\times n}$ is a rotation matrix. Then if $Q$ is an isometry, we require:
\begin{equation}
    \begin{split}
        \mathbf{x}^\top \mathbf{y} &= (Q\mathbf{x})^\top (Q\mathbf{y})\\
        &=\mathbf{x}^\top (Q^\top Q) \mathbf{y}
    \end{split}
\end{equation}
or $Q^\top Q = I = Q^\top Q$. Preservation of orientation (equivalently, handedness) means $\det Q > 0$. We take the determinant of the identity and find $\det(Q^\top Q)=(\det(Q))^2=1$, which means $\det Q = \pm 1$. Hence, $\det Q = +1$ for $SO(n)$.
\section{Proof of Equivariances for Temporal Attention Layer}\label{B}

\subsection{Position Component: $E(n)$-Equivariance}

\begin{proof}
    Consider an isometry of $E(n)$ defined by $\xib\mapsto A\xib + \mathbf{b}$ for $A\in\mathbb{R}^{n\times n}$ an orthogonal rotation matrix and $\mathbf{b}\in\mathbb{R}^n$ a translation vector. Then
    \begin{equation}
        \begin{split}
            ||\xib_i(t)-\xib_i(s)||^2&=||A\xib_i(t) + \mathbf{b} - A\xib_i(s) - \mathbf{b} ||^2\\
            &=||A(\xib_i(t)-\xib_i(s))||^2\\
            &=(\xib_i(t)-\xib(s))^\top A^\top A(\xib_i(t)-\xib_i(s))\\
            &=||\xib_i(t)-\xib_i(s)||^2,
        \end{split}
    \end{equation}
where the last line follows by the orthogonality of $A$. Hence, $\beta_i(t,s)=\exp(||\xib_i(t)-\xib_i(s)||^2)/\sum_{s'=1}^L\exp(||\xib_i(t)-\xib_i(s')||^2)$ is invariant under isometries of $\mathbb{R}^n$. Thus, applying the isometry to the layer 
\begin{equation}
\tilde{\xib}_i(t)=\xib_i(t)+B\sum_{\substack{s=1 \\ s \ne t}}^L \beta_i(t,s)(\xib_i(s)-\xib_i(t))
\end{equation}
yields:
\begin{equation}
    \begin{split}
        &A\xib_i(t) + \mathbf{b} + B\sum_{\substack{s=1 \\ s \ne t}}^L \beta_i(t,s) A(\xib_i(s)-\xib_i(t))\\
        &=A\left(\xib_i(t) + B\sum_{\substack{s=1 \\ s \ne t}}^L \beta_i(t,s) (\xib_i(s)-\xib_i(t))\right) + \mathbf{b}\\
        &=A\tilde{\xib}_i(t) + \mathbf{b}.
    \end{split}
\end{equation}
It follows that the position component of ETAL is $E(n)$-equivariant.
\end{proof}

\subsection{Velocity Component: $SO(n)$-Equivariance}

\begin{proof}
    Consider an orthogonal rotation matrix $A\in\mathbb{R}^{n\times n}$ of $SO(n)$ with action $\omegab\mapsto A\omegab$. Clearly,
    \begin{equation}
        \begin{split}
            (A\omegab_i(t))^\top (A\omegab_i(s))&= \omegab_i(t)^\top A^\top A\omegab_i(s)\\
            &=\omegab_i(t)^\top \omegab_i(s).
        \end{split}
    \end{equation}

    Thus, the coefficient $\gamma_i(t,s)=\omegab_i(t)^\top \omegab_i(s)/\sum_{s'=1}^L \exp(\omegab_i(t)^\top \omegab_i(s'))$ is invariant under rotations of $SO(n)$.

    Applying the rotation to the attention layer
    \begin{equation}
        \tilde{\omegab}_i(t)=\sum_{s=1}^L \gamma_i(t,s)\omegab_i(s)
    \end{equation}
    yields:
    \begin{equation}
        \begin{split}
            \sum_{s=1}^L \gamma_i(t,s)A \omegab_i(s)&=A\sum_{s=1}^L \gamma_i(t,s)\omegab_i(s)\\
            &=A\tilde{\omegab}_i(t),
        \end{split}
    \end{equation}
as asserted. Hence, the velocity component of ETAL is $SO(n)$-equivariant.
\end{proof}

\section{Tensorization of Equivariant Temporal Attention Layer (ETAL)}\label{C}

\subsection{Temporal Attention Feature Component}
Define the matrix of spatial graph representations for features across time:

\begin{equation}
    \theta_i^{[1:L]} = \begin{bmatrix}
           \thetab_i(1)^\top \\
           \vdots \\
           \thetab_i(L)^\top
         \end{bmatrix}\in\mathbb{R}^{L\times d}.
\end{equation}
We compute the value vectors $v_i^{[1:L]}=\theta_i^{[1:L]} V\in\mathbb{R}^{L\times d}$, key vectors $k_i^{[1:L]}=\theta_i^{[1:L]} K\in\mathbb{R}^{L\times d}$, and query vectors $q_i^{[1:L]}=\theta_i^{[1:L]} Q\in\mathbb{R}^{L\times d}$ for $V, K, Q\in\mathbb{R}^{d\times d}$. Thus, to compute the attention layer for the graph features in Equation \ref{feature_update}, we first compute the weights:
\begin{equation}
    \alpha_i = \text{softmax}\left(\frac{\theta_i^{[1:L]}QK^\top {\theta_i^{[1:L]}}^\top}{\sqrt{d}}\right)\in\mathbb{R}^{L\times L}.
\end{equation}

Then we apply the weights as follows:
\begin{equation}
    \tilde{\theta}_i^{[1:L]}=\alpha_i v_i^{[1:L]} \in\mathbb{R}^{L\times d}.
\end{equation}

We would like to tensorize this computation for all nodes $i=1,\dots,N$. Let $\theta^{[1:L]}=(\theta_1^{[1:L]},\dots, \theta_N^{[1:L]})\in\R^{N\times L\times d}$. Likewise, let $k^{[1:L]}=\theta^{[1:L]}K\in\R^{N\times L\times d}$, $q^{[1:L]}=\theta^{[1:L]}Q\in\R^{N\times L\times d}$, and $v^{[1:L]}=\theta^{[1:L]}V\in\R^{N\times L\times d}$. In a slight abuse of notation,
\begin{equation}
    \alpha=\text{softmax}\left(\frac{q^{[1:L]}{k^{[1:L]}}^\top}{\sqrt{d}}\right)\in\R^{N\times L\times L}
\end{equation}
where the transpose ${k^{[1:L]}}^\top\in\R^{N\times d \times L}$ interchanges the last two dimensions, the tensor multiplication $q^{[1:L]}{k^{[1:L]}}^\top$ is with respect to the last two dimensions, and the softmax is computed over the last dimension. Then
\begin{equation}
    \tilde{\theta}^{[1:L]}=\alpha v^{[1:L]} \in\R^{N\times L
    \times d},
\end{equation} where the tensor multiplication is over the last two dimensions.
\subsection{Temporal Attention Position Component}

Define the matrix of spatial graph representations for positions and its corresponding temporally-transformed matrix, across time:

\begin{equation}
    \xi_i^{[1:L]}=\begin{bmatrix}
           \xib_i(1)^\top \\
           \vdots \\
           \xib_i(L)^\top
         \end{bmatrix}\in\mathbb{R}^{L\times n},
        \tilde{\xi}_i^{[1:L]}=\begin{bmatrix}
           \tilde{\xib}_i(1)^\top \\
           \vdots \\
           \tilde{\xib}_i(L)^\top
         \end{bmatrix}\in\mathbb{R}^{L\times n}.
\end{equation} 
Similarly, construct the node-wise attention matrix: 
\begin{equation}
    \beta_i = \begin{bmatrix}
             \beta_i(1,1) & \hdots & \beta_i(1,L)\\
             \vdots & \ddots & \vdots \\
             \beta_i(L,1) & \hdots & \beta_i(L,L)
         \end{bmatrix},
\end{equation} where $\beta_i(t,s)$ is defined in Equation \ref{beta}. Then we can write the update in Equation \ref{position_update} as:
\begin{equation}
    \begin{split}
        \tilde{\xi}_i^{[1:L]} &= \xi_i^{[1:L]} + B\left((1_{L\times L} - I_{L\times L}) \odot \beta_i\right) \xi_i^{[1:L]} \\
        &- B\left(((1_{L\times L} - I_{L\times L})\odot \beta_i) 1_{L\times 1}\right)\odot \xi_i^{[1:L]},
    \end{split}
\end{equation}
where $\odot$ is the Hadamard product. Define the difference matrix:

\begin{equation}
    \begin{split}
        D_i:&=\begin{bmatrix}
        \xib_i(1)-\xib_i(1) & \hdots & \xib_i(1)-\xib_i(L) \\
        \vdots & \ddots & \vdots \\
        \xib_i(L)-\xib_i(1) & \hdots & \xib_i(L)-\xib_i(L)
    \end{bmatrix}
    \\
    &=\begin{bmatrix}
        \xib_i(1) & \hdots & \xib_i(1) \\
        \vdots & \ddots & \vdots \\
        \xib_i(L) & \hdots & \xib_i(L)
    \end{bmatrix}
    -
    \begin{bmatrix}
        \xib_i(1) & \hdots & \xib_i(L) \\
        \vdots & \ddots & \vdots \\
        \xib_i(1) & \hdots & \xib_i(L)
    \end{bmatrix}.
    \end{split}
\end{equation} 
Then we can compute the weights as:
\begin{equation}
    \beta_i=\text{softmax}\left(\frac{-D_i^\top \odot D_i}{\sqrt{n}}\right) \in\mathbb{R}^{L\times L}.
\end{equation}

Again, we seek to find a differentiable expression that tensorizes the attention layer for all nodes. We stack $\tilde{\xi}_i^{[1:L]}$ for $i=1,\dots, N$ into a tensor:

\begin{equation}
    \begin{split}
        \tilde{\xi}^{[1:L]}&=\begin{bmatrix}
        \tilde{\xi}_1^{[1:L]}\\
        \vdots \\
        \tilde{\xi}_N^{[1:L]}
    \end{bmatrix}\\
    &=\begin{bmatrix}
        \xi_1^{[1:L]}\\
        \vdots \\
        \xi_N^{[1:L]}
    \end{bmatrix} + B \begin{bmatrix}
        ((1_{L\times L} - I_{L\times L})\odot \beta_1)\xi_1^{[1:L]}\\
        \vdots \\
        ((1_{L\times L} - I_{L\times L})\odot \beta_N)\xi_N^{[1:L]}
    \end{bmatrix}
    - B \begin{bmatrix}
        (((1_{L\times L} - I_{L\times L})\odot \beta_1)1_{L\times 1}) \odot \xi_1^{[1:L]}\\
        \vdots \\
        (((1_{L\times L} - I_{L\times L})\odot \beta_N)1_{L\times 1}) \odot \xi_N^{[1:L]}
    \end{bmatrix}\\
    &=\begin{bmatrix}
        \xi_1^{[1:L]}\\
        \vdots \\
        \xi_N^{[1:L]}
    \end{bmatrix}
    + B \begin{bmatrix}
        (1_{L\times L} - I_{L\times L})\odot \beta_1\\
        \vdots \\
        (1_{L\times L} - I_{L\times L})\odot \beta_N
    \end{bmatrix} \odot 
    \begin{bmatrix}
        \xi_1^{[1:L]}\\
        \vdots \\
        \xi_N^{[1:L]}
    \end{bmatrix}
    - B \begin{bmatrix}
        (((1_{L\times L} - I_{L\times L})\odot \beta_1)1_{L\times 1}) \odot \xi_1^{[1:L]}\\
        \vdots \\
        (((1_{L\times L} - I_{L\times L})\odot \beta_N)1_{L\times 1}) \odot \xi_N^{[1:L]}
    \end{bmatrix},
    \end{split}
\end{equation}
which is a differentiable function with respect to the $\xib_i(1), \dots, \xib_i(L)$ for $i=1,\dots, N$, where $\tilde{\xi}^{[1:L]}\in\R^{N\times L\times n}$.

\subsection{Temporal Attention Velocity Component}

Define the matrix of spatial graph representations for velocities, across time:

\begin{equation}
    \omega_i^{[1:L]} = \begin{bmatrix}
           \omegab_i(1)^\top \\
           \vdots \\
           \omegab_i(L)^\top
         \end{bmatrix}\in\mathbb{R}^{L\times n}.
\end{equation}

We compute the weights as:
\begin{equation}
    \gamma_i = \text{softmax}\left(\frac{\omega_i^{[1:L]}{\omega_i^{[1:L]}}^\top}{\sqrt{n}}\right) \in \mathbb{R}^{L\times L}.
\end{equation}
Then we apply the weights to obtain the layer transform:
\begin{equation}
    \tilde{\omega}_i^{[1:L]} = \gamma_i \omega_i^{[1:L]}\in\mathbb{R}^{L\times n}.
\end{equation}

We would like to find a tensorized expression that includes all nodes $i=1,\dots,N$. Let $\omega^{[1:L]}=(\omega_1^{[1:L]},\dots, \omega_N^{[1:L]})\in\R^{N\times L\times n}$. As before, in a slight abuse of notation,
\begin{equation}
    \gamma=\text{softmax}\left(\frac{{\omega}^{[1:L]}{{\omega}^{[1:L]}}^\top}{\sqrt{n}}\right)\in\R^{N\times L\times L},
\end{equation}
where the transpose ${{\omega}^{[1:L]}}^\top\in\R^{N\times n \times L}$ interchanges the last two dimensions, the tensor multiplication ${\omega}^{[1:L]}{{\omega}^{[1:L]}}^\top$ is with respect to the last two dimensions, and the softmax is computed over the last dimension. Hence,
\begin{equation}
    \tilde{\omega}^{[1:L]}=\gamma \omega^{[1:L]} \in\R^{N\times L
    \times n},
\end{equation} where the tensor multiplication is over the last two dimensions.

\subsection{Temporal Attention Adjacency Component}

Let $A^{[1:L]}=(A(1),\dots,A(L))\in\R^{L\times N\times N}$ be the tensor of all adjacency matrices over time. Let $k^{[1:L]}=A^{[1:L]}K$, $q^{[1:L]}=A^{[1:L]}Q$, $v^{[1:L]}=A^{[1:L]}V$ where $K, Q, V\in\R^{N\times N}$. With a slight abuse of notation, let
\begin{equation}
    \pi = \text{softmax}\left(\frac{q^{[1:L]}{k^{[1:L]}}^\top}{\sqrt{N}}\right) \in\R^{L\times N\times N},
\end{equation}
where the transpose ${k^{[1:L]}}^\top\in \R^{L\times N\times N}$ interchanges the last two dimensions, the tensor multiplication $q^{[1:L]}{k^{[1:L]}}^\top$ is with respect to the last two dimensions, and the softmax is computed over the last dimension as usual. 

Thus,
\begin{equation}
    \tilde{A}^{[1:L]}=\pi v^{[1:L]} \in \R^{L\times N\times N}.
\end{equation}

\section{Positional Encoding \& Layer Normalization}\label{D}

We use sinusoidal positional encodings, as defined in \cite{vaswani}. That is, we define the positional encodings $W^{[1:L]}\in\R^{L\times N\times d}$, $X^{[1:L]}\in\R^{L\times N\times n}$, $Y^{[1:L]}\in\R^{L\times N\times n}$, and $Z^{[1:L]}\in\R^{L\times N^2}$, used in Algorithm \ref{alg:attention}, as:
\begin{equation}
    \begin{split}
        W_{i, \cdot, 2j}&=\sin\left(\frac{i}{\kappa^{2j/d}}\right), W_{i, \cdot, 2i+1}=\cos\left(\frac{i}{\kappa^{2j/d}}\right),\\
        X_{i, \cdot, 2j}&=\sin\left(\frac{i}{\kappa^{2j/n}}\right), X_{i, \cdot, 2i+1}=\cos\left(\frac{i}{\kappa^{2j/n}}\right),\\
        Y_{i, \cdot, 2j}&=\sin\left(\frac{i}{\kappa^{2j/n}}\right), Y_{i, \cdot, 2i+1}=\cos\left(\frac{i}{\kappa^{2j/n}}\right),\\
        Z_{i, 2j}&=\sin\left(\frac{i}{\kappa^{2j/N^2}}\right), Z_{i, 2i+1}=\cos\left(\frac{i}{\kappa^{2j/N^2}}\right).
    \end{split}
\end{equation}

We compress the adjacency matrix positional encoding as $Z^{[1:L]}\in\R^{L\times N^2}$ because we treat the $N^2$ entries of the adjacency matrix as belonging to a vectorspace, to which we apply a positional encoding.

Layer normalization in Algorithm \ref{alg:attention} is defined as follows:
\begin{equation}
    \begin{split}
        &\hat{\mu}^{[1:L]} = \begin{bmatrix}
        \frac{1}{d}\sum_{j=1}^d \theta_{1, j}^{[1:L]} \\
        \vdots \\
        \frac{1}{d}\sum_{j=1}^d \theta_{LN, j}^{[1:L]}
    \end{bmatrix}\in\R^{L\times N\times 1},\\ 
    &\hat{\sigma}^{[1:L]} = \begin{bmatrix}
        \sqrt{\frac{1}{d}\sum_{j=1}^d \left(\theta_{1,j}^{[1:L]}-\hat{\mu}_1^{[1:L]}\right)^2} \\
        \vdots \\
        \sqrt{\frac{1}{d}\sum_{j=1}^d \left(\theta_{LN,j}^{[1:L]}-\hat{\mu}_{LN}^{[1:L]}\right)^2}
    \end{bmatrix}\in\R^{L\times N\times 1},\\
&LN(\theta^{[1:L]})=\frac{\theta^{[1:L]} - \hat{\mu}^{[1:L]}}{\hat{\sigma}^{[1:L]}},
    \end{split}
\end{equation}
so we broadcast $\hat{\mu}, \hat{\sigma}$ across the second $d$ dimensions of $\theta^{[1:L]}$. Likewise, for the adjacency matrix, we define:
\begin{equation}
    \begin{split}
        &\hat{\mu}^{[1:L]} = \begin{bmatrix}
        \frac{1}{N^2}\sum_{j=1}^{N^2} \theta_{1, j}^{[1:L]} \\
        \vdots \\
        \frac{1}{N^2}\sum_{j=1}^{N^2} \theta_{L, j}^{[1:L]}
    \end{bmatrix}\in\R^{L\times 1},\\ 
    &\hat{\sigma}^{[1:L]} = \begin{bmatrix}
        \sqrt{\frac{1}{N^2}\sum_{j=1}^{N^2} \left(\theta_{1,j}^{[1:L]}-\hat{\mu}_1^{[1:L]}\right)^2} \\
        \vdots \\
        \sqrt{\frac{1}{N^2}\sum_{j=1}^{N^2} \left(\theta_{L,j}^{[1:L]}-\hat{\mu}_{L}^{[1:L]}\right)^2}
    \end{bmatrix}\in\R^{L\times 1},\\
&LN(A^{[1:L]})=\frac{A^{[1:L]} - \hat{\mu}^{[1:L]}}{\hat{\sigma}^{[1:L]}},
    \end{split}
\end{equation}
where we view $A^{[1:L]}\in\R^{L\times N^2}$.

\section{Hyper-parameter Settings \& Implementation Details}\label{E}

\subsection{SET \& Scaling Results}

After a hyper-parameter optimization for SET, we found the following optimal settings. Namely, we used an Adam optimizer with an initial learning rate of $4.45\times 10^{-5}$, a batch size of $100$, dropout of $0.1$, $2$ EGNN layers with SiLU activations, $3$ vertical stacking layers with ReLU activations, hidden dimension of $128$, spatial and temporal attention, and equivariance. The following hyper-parameters were inactive: weight decay, temporal adjacency, positional encoding and causal attention. We chose $\alpha=1$ in Equation \ref{objective} and $B=0.5$ in Equation \ref{position_update}.

The results in tables \ref{tab:n20_metrics} and \ref{tab:n30_metrics} show the effects of scaling the dataset. SET is the best-performing model across all datasets.

\begin{table}[H]
\centering
\renewcommand{\arraystretch}{1.5}
\begin{tabular}{lcc}
\hline
\textbf{Model} & \textbf{Params} & \textbf{Test MSE} \\
\hline
SET & 796,058 & \textbf{5.72e-11} \\
LSTM & 1,269,833 & 1.02e-04 \\
EGNN & 100,612 & 5.34e-05 \\
MLP & 67,718 & 3.77e-06 \\
Linear & 3 & 4.48 \\
\hline
\end{tabular}
\vspace{0.5cm}
\caption{Performance metrics for optimized models with $N = 20$.}
\label{tab:n20_metrics}
\end{table}

\begin{table}[H]
\centering
\renewcommand{\arraystretch}{1.5}
\begin{tabular}{lcc}
\hline
\textbf{Model} & \textbf{Params} & \textbf{Test MSE} \\
\hline
SET & 796,058 & \textbf{1.20e-10} \\
LSTM & 1,859,513 & 1.95e-07 \\
EGNN & 100,612 & 5.88e-05 \\
MLP & 67,718 & 3.99e-06 \\
Linear & 3 & 5.19 \\
\hline
\end{tabular}
\vspace{0.5cm}
\caption{Performance metrics for optimized models with $N = 30$.}
\label{tab:n30_metrics}
\end{table}

\subsection{EGNN, LSTM, MLP, Linear Baselines}

For the EGNN baseline, we apply the same methodology outlined in Algorithm \ref{alg:attention}, whereby we apply MLPs $\phi_e,\phi_v,\phi_x,\phi_h$ across the time dimension:
\begin{equation}
    \begin{split}
        \h_i^{(l+1)}(1)&=\phi_h(\h_i^{(l)}(1), \m_i(1))\in\R^d\\
        &\vdots \\
        \h_i^{(l+1)}(L)&=\phi_h(\h_i^{(l)}(L), \m_i(L))\in\R^d,
    \end{split}
\end{equation}
for $l=1,\dots, K$. We used an Adam optimizer without weight decay and with an initial learning rate of $3.98\times 10^{-5}$, batch size of $32$, hidden dimension of $64$, $3$ EGNN layers, and residual connections.

For the LSTM baseline, we found using only temporal attention was optimal. That is, we did not pre-process the graph data with an EGNN; we simply applied a vanilla LSTM to the graph data. Each token embedding $\mathbf{z}(t)\in\R^{N\times(d + 2n + 2N -2)}$ comprises of the following data: 
\begin{equation}
    \begin{split}
        &\h_1(t),\dots,\h_N(t) \in\mathbb{R}^{N\times d}\\
        &\x_1(t),\dots,\x_N(t)\in\mathbb{R}^{N\times n}\\
        &\vel_1(t),\dots,\vel_N(t)\in\mathbb{R}^{N\times n}\\
        &\textbf{e}_{0,1}(t),\dots,\textbf{e}_{0,N}(t),\textbf{e}_{1,0}(t),\dots, \textbf{e}_{1,N}(t),\\
        &\dots,\textbf{e}_{N,0}(t), \dots, \textbf{e}_{N, N-1}(t)\in\mathbb{R}^{2}.
    \end{split}
\end{equation}

We used Adam with an initial learning rate of $2.47\times 10^{-5}$, batch size of $32$, hidden dimension of $512$, $2$ EGNN layers, $3$ temporal stacking layers, and dropout of $0.1$. We omitted the use of weight decay, temporal adjacency, and recurrence. The MLP uses $5$ hidden layers with hidden dimension $128$. It only takes a tensor of positions $x\in\R^{L\times N\times n}$ and velocities $v\in\R^{L\times N\times n}$ as inputs. It is trained using an Adam optimizer with a learning rate of $1.75\times 10^{-5}$ and a batch size of $100$. The linear dynamics model simply predicts position $x(t)=x(t-1)+ \alpha v(t-1)$ and velocity $v(t) = \beta v(t-1) + \gamma$, learning $\alpha, \beta, \gamma$ using an Adam optimizer with initial learning rate of $2.73\times 10^{-5}$, weight decay of $1\times 10^{-6}$, and batch size of $100$.
\end{document}